\documentclass[10pt,twocolumn,letterpaper]{article}
\pdfoutput=1

\usepackage{cvpr}
\usepackage{times}
\usepackage{epsfig}
\usepackage{graphicx}
\usepackage{amsmath}
\usepackage{amssymb}

\usepackage{multirow}
\usepackage{graphicx}
\usepackage{subfigure}
\usepackage{colortbl}
\usepackage{algorithm}
\usepackage{algorithmic}
\usepackage{booktabs}
\usepackage{authblk}

\usepackage[breaklinks=true,bookmarks=false]{hyperref}

\cvprfinalcopy 

\def\cvprPaperID{****} 

\begin{document}
	
	\title{Balanced Knowledge Distillation for Long-tailed Learning}
	


	\author[1,2]{Shaoyu Zhang}
	\author[1,2]{Chen Chen \thanks{Corresponding author}}
	\author[3]{Xiyuan Hu}
	\author[1,2]{Silong Peng}
	\affil[1]{Institute of Automation, Chinese Academy of Sciences}
	\affil[2]{University of Chinese Academy of Sciences}
	\affil[3]{Nanjing University of Science and Technology}
	
	\maketitle
	
	\begin{abstract}
		Deep models trained on long-tailed datasets exhibit unsatisfactory 
		performance on tail classes. Existing methods usually modify the 
		classification loss to increase the learning focus on tail classes, 
		which unexpectedly sacrifice the performance on head classes. 
		In fact, this scheme leads to a contradiction between the two 
		goals of long-tailed learning, i.e., learning generalizable 
		representations and facilitating learning for tail classes. 
		In this work, we explore knowledge distillation in long-tailed 
		scenarios and propose a novel distillation framework, named 
		{\itshape Balanced Knowledge Distillation (BKD)}, to disentangle the 
		contradiction between the two goals and achieve both simultaneously. 
		Specifically, given a vanilla teacher model, we train the 
		student model by minimizing the combination of an instance-balanced 
		classification loss and a class-balanced distillation loss. 
		The former benefits from the sample diversity and learns 
		generalizable representation, while the latter considers the 
		class priors and facilitates learning mainly for tail classes. 
		The student model trained with BKD obtains significant 
		performance gain even compared with its teacher model. 
		We conduct extensive experiments on several long-tailed benchmark 
		datasets and demonstrate that the proposed BKD is an effective knowledge distillation framework
		in long-tailed scenarios, as well as a new state-of-the-art method for long-tailed learning. Code is available at 
		\url{https://github.com/EricZsy/BalancedKnowledgeDistillation}.
	\end{abstract}
	
	
	\section{Introduction}
	
	\begin{figure}[h]
		\centering
		\includegraphics[width=\linewidth]{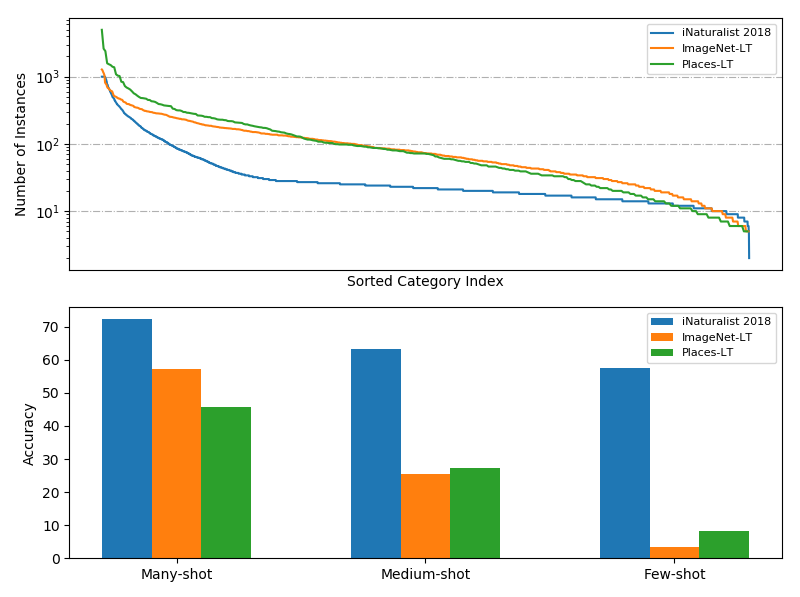}
		\caption{Illustration of the relation between class capacity and classification accuracy on
			Many/Medium/Few-shot subsets.
			Top: Number of instances per class in three long-tailed dataset, 
			sorted in descending order. Bottom: Per-subset accuracy
			of a model trained with vanilla cross-entropy loss.}
		\label{instance}
	\end{figure}
	
	Recent advances in visual recognition \cite{he2016deep, long2015fully, szegedy2015going} are mainly driven by the use of large-scale 
	datasets, such as ImageNet ILSVRC 2012 \cite{deng2009imagenet, russakovsky2015imagenet} and MS COCO \cite{lin2014microsoft}. Such datasets are 
	often carefully collected, with roughly balanced quantities in each category.
	However, in practical scenarios, data 
	tends to exhibit long-tailed distribution \cite{reed2001pareto, horn2017devil}, wherein a few classes (head classes)
	have a significantly larger number of instances than other classes (tail classes). 
	This uneven distribution affects both convergence during the training phase 
	and generalization on the test set \cite{buda2018systematic}. When dealing with such 
	imbalanced data, deep models tend to bias towards head classes, 
	resulting in performance drop on tail classes \cite{he2009learning, japkowicz2002class, li2020overcoming}.
	Figure \ref{instance} illustrates the relation between the data distribution
	and the classification accuracy on Many/Medium/Few-shot subsets. As we can see,
	when the data distribution is highly skewed, the model trained with vanilla cross-entropy 
	loss exhibits great performance gaps between the three subsets.
	While most of the samples from the head classes are classified correctly,
	the classification accuracy of tail classes is even lower than $10\%$ in some datasets, 
	e.g., ImageNet-LT \cite{liu2019large} and Places-LT \cite{liu2019large}.
	
	The goals of long-tailed learning is two-fold: learning generalizable representations
	and facilitating learning for tail classes. 
	In the literature, one of the most common practices to facilitate learning for tail
	classes is to re-balance the class distribution, 
	either by re-sampling the examples 
	\cite{drummond2003c4, byrd2019effect, chawla2002smote, kubat1997addressing}
	or re-weighting the 
	classification loss \cite{huang2016learning, khan2017cost, cui2019class, tan2020equalization}. 
	Essentially, these methods aim to seek trade-offs between the 
	accuracies of the head classes and the tail classes and 
	improve overall performance. Although effective, such re-balancing strategies
	will encourage excessive focus on tail classes and damage 
	the overall representation learning to some extent \cite{cao2019learning, zhou2020bbn}. To solve this problem, 
	recent methods propose two-phase \cite{cao2019learning, kang2019decoupling} or 
	bilateral-branch \cite{zhou2020bbn}
	training framework to decouple the 
	learning procedure into representation learning and classification.
	It is noteworthy that the decoupled learning scheme 
	exhibits an interesting phenomena that 
	the vanilla instance-balanced cross-entropy loss gives the most generalizable
	representations.
	
	Most aforementioned works make modifications directly on the classification loss. 
	However, this scheme is inconsistent with the principle of cross-entropy loss and 
	naturally brings along with a contradiction 
	between learning generalizable representations
	and facilitating learning for tail classes. 
	We take the most representative re-weighting strategies \cite{cui2019class} for example and
	explain where the contradiction comes from by analyzing the effect of gradient.
	
	Orthogonally to the previous works, we disentangle the contradiction
	between the two goals from the perspective of teacher-student learning 
	\cite{hinton2015distilling, yim2017gift, wang2021knowledge}. 
	Specifically, if trained on imbalanced data, the student tries the best to
	learn high-quality representations while the teacher should facilitate 
	learning with focus on tail classes.
	Motivated by this, we explore knowledge distillation in long-tailed
	scenarios and propose a simple yet effective distillation framework, 
	named {\itshape balanced knowledge distillation (BKD)}, to alleviate the long-tailed problem. 
	We first train a vanilla teacher model with the same size as the student model
	by minimizing cross-entropy loss. Then the student model is trained by 
	minimizing the combination of 
	instance-balanced classification loss and class-balanced distillation loss.
	Thus, the learning procedure of student model is decoupled into two tasks, where
	each performs its own duty for learning representations and
	facilitating learning for tail classes, respectively. 
	
	In experiments, we observe that the student model trained with BKD even
	outperforms the teacher model by a large margin, e.g., 10 points accuracy
	gain on long-tailed CIFAR-10 \cite{krizhevsky2009learning}. Furthermore, we conduct extensive experiments
	on several large-scale real-world benchmarks including iNaturalist 2018 \cite{van2018inaturalist}, 
	ImageNet-LT \cite{liu2019large} and Places-LT \cite{liu2019large}.
	Experimental results justify that our BKD framework is able to achieve state-of-the-art
	performance.
	
	Our key contributions can be summarized as follows: 
	\begin{itemize}
		\item We analyze the underlying cause of the failure of re-weighting methods 
		from the perspective of gradient. Accordingly, we discuss our motivation of
		disentangling the contradiction between the two goals of 
		learning generalizable representations 
		and facilitating learning for tail classes.
		\item We fill the research gap of knowledge distillation on imbalanced data 
		and propose {\itshape balanced knowledge distillation (BKD)} as a more effective distillation
		framework in long-tailed scenarios. When we compare the student model trained with BKD and the teacher model
		with the same size as the student model, we observe significant performance gain
		from the former.
		\item We conduct extensive experiments and 
		demonstrate that our BKD framework significantly 
		improves the classification performance of the student model
		and achieves state-of-the-art performance on several long-tailed benchmarks.
	\end{itemize}

	\section{Related Works}
	
	\subsection{Knowledge Distillation}
	
	The idea of training a lightweight 
	student model to mimic a larger but better performing teacher model can 
	be traced back to \cite{bucilu2006model, 10.5555/2969033.2969123}. 
	Hinton et al. \cite{hinton2015distilling} propose to distill the knowledge 
	of predicted distribution from the teacher model into the student model, 
	which is widely known as {\itshape knowledge distillation}. Apart from the final 
	prediction \cite{meng2019conditional}, other types of knowledge, 
	like the intermediate representations \cite{romero2014fitnets, komodakis2017paying, heo2019knowledge}
	or the relationships between different layers \cite{yim2017gift} or 
	data samples \cite{you2017learning, park2019relational, liu2019knowledge},
	can also be used to guide the learning of the student model. As a 
	special case, knowledge distillation from a model to another 
	of identical architecture \cite{furlanello2018born, yim2017gift} is explored. 
	However, to our knowledge, how to effectively 
	distill the knowledge in large-scale long-tailed scenarios is still
	under-explored. We progress this line of works under two challenges: 
	1) the teacher and the student have the same architecture; 
	2) the training data is imbalanced.
	
	\subsection{Long-tailed Learning}
	
	To alleviate the challenge of long-tailed learning, most of pioneer 
	works have been proposed from three aspects:
	
	\noindent
	{\bfseries Data.}\quad Re-sampling 
	methods \cite{byrd2019effect, buda2018systematic, shen2016relay, chawla2002smote, he2009learning} 
	created a roughly balanced distribution 
	by either over-sampling or under-sampling. Over-sampling \cite{shen2016relay, byrd2019effect} repeatedly 
	samples training examples from the minority classes, the downside of 
	which is the high potential risk of overfitting. To overcome this 
	issue, SMOTE \cite{chawla2002smote, han2005borderline} is proposed to 
	augment synthetic data created by 
	interpolating neighboring data points. As opposed to over-sampling, 
	under-sampling \cite{he2009learning, japkowicz2002class, buda2018systematic} 
	randomly discards examples from the majority classes. 
	When the imbalance is extreme, under-sampling may lose valuable 
	information in majority classes.
	
	\noindent
	{\bfseries Optimization objective.}\quad The key to this 
	line of work 
	\cite{huang2016learning, lin2017focal, cui2019class, sarafianos2018deep, zhang2017range, NEURIPS2020_2ba61cc3} 
	is to adjust learning focus by modifying objective 
	functions. Cost-sensitive re-weighting 
	\cite{elkan2001foundations, ting2000comparative, khan2017cost} assigns different weights 
	to the classification loss terms corresponding to different classes \cite{cui2019class, khan2017cost, cao2019learning}
	or different samples \cite{lin2017focal, khan2019striking, li2019gradient}. 
	The traditional strategy re-weights classes 
	proportionally to the inverse of their frequency of samples \cite{huang2016learning}. 
	Taking data overlap into consideration, Cui et al. \cite{cui2019class} design a 
	class-balanced cross-entropy loss based on effective number of samples 
	in each class. Another important work \cite{cao2019learning} adds a class-wise 
	margin in to the cross-entropy loss motivated by minimizing a 
	margin-based generalization bound. Recently, Menon et al. \cite{menon2020long} revisited 
	the method of logit adjustment and proposed a general pair-wise margin loss 
	with several previous methods \cite{cao2019learning, tan2020equalization} as its special cases.
	
	Although effective, re-weighting on classification loss has a negative 
	effect on representation learning \cite{zhou2020bbn}. Motivated by this observation, 
	two-stage \cite{cao2019learning, kang2019decoupling} and two-branch \cite{zhou2020bbn} methods are proposed to 
	take care of both representation learning and classifier learning.
	
	\noindent
	{\bfseries Meta- and dark-knowledge.}\quad Many approaches 
	\cite{jamal2020rethinking, ren2018learning, xiang2020learning} design additional 
	modules or meta-network to transfer knowledge, e.g., 
	distributions \cite{yin2019feature, liu2020deep}, memory features \cite{liu2019large} or meta-knowledge \cite{wang2017learning}, from head to tail 
	classes. However, this line of work is usually non-trivial and 
	more computationally expensive. Beyond that, as a popular technique 
	of transferring knowledge, knowledge distillation attracts 
	attention in the field of long-tailed learning. Recently, 
	Learning From Multiple Experts (LFME) \cite{xiang2020learning} has been proposed as a multi-teacher 
	framework, in which each teacher learns from a relatively balanced 
	subset. LFME also employs self-paced expert selection and 
	curriculum learning strategy.
	
	\noindent
	{\itshape Key differences.}\quad Our work differs from LFME \cite{xiang2020learning} in several ways. 
	On one hand, LFME needs to split training data into groups and train 
	multiple teacher models, while our method only trains a single teacher model 
	on the original data with vanilla cross-entropy loss. On the other hand, 
	LFME involves complex adaptive learning schedules at model level and 
	instance level, while our method simply modifies the knowledge 
	distillation loss to control the focus of distillation process. Indeed, 
	without multiple teachers and complex learning schedules, our method 
	outperforms LFME on all the long-tailed benchmarks we report by a large margin.
	
	\section{Insight and Motivation}
	
	{\bfseries Notation.} Consider a classification problem on long-tailed 
	training data. Let 
	\begin{math}
		x \in R^d
	\end{math} and
	\begin{math}
		y \in \{1, \dots, C\}
	\end{math} denote a data point and its label, respectively. Due to the 
	imbalanced distribution, the number of training examples in each class
	\begin{math}
		n_i
	\end{math} is highly imbalanced. Without loss of generality, we sort the 
	classes in descending order of frequency so that
	\begin{math}
		n_1>\dots>n_C
	\end{math}. Our goal is to learn a model
	\begin{math}
		f: R^d \to R^c
	\end{math} that estimates the conditional probability
	\begin{math}
		p_i=softmax(z_i)
	\end{math} from the network output
	\begin{math}
		z=[z_1,\dots,z_C]^T
	\end{math}.
	
	\subsection{A Closer Look at Re-weighting Methods}\label{rw}
	
	Given a dataset, the most straightforward method 
	minimizes the misclassification error by minimizing the following softmax cross-entropy loss
	\begin{equation}\label{CELoss}
		L_{CE}=-\sum_i y_i\log p_i
	\end{equation}
	On the basis of cross-entropy loss, the re-weighting methods typically 
	assign weights for different classes or even different samples.
	However, a possible side effect of re-weighting is that the model tends to 
	overfit on tail classes and suffer a performance drop on head classes. 
	Cao et al. \cite{cao2019learning} and Zhou et al. \cite{zhou2020bbn} 
	experimentally show that the re-weighting methods have an adverse 
	effect on representation learning. We start by analyzing the 
	underlying cause of such problems from the perspective of gradient.
	
	The key of re-weighting is to balance the classification loss by 
	a weight vector
	\begin{math}
		\omega=[\omega_1,\dots,\omega_C]^T
	\end{math}.
	\begin{math}
		\omega_i
	\end{math} is typically a decreasing transform of
	\begin{math}
		n_i
	\end{math}. 
	As a representative work of re-weighting methods, 
	Cui et al. \cite{cui2019class} formulate the weight factor inversely proportional to the effective 
	number of samples for class $i$
	\begin{equation} \label{class-balanced term}
		\omega_i = \frac{1-\beta}{1-\beta^{n_i}},
	\end{equation}
	where $\beta \in (0,1)$ is a hyperparameter to 
	adjust the class balanced term.
	In this setting, we have
	\begin{math}
		\omega_1 < \dots <\omega_C
	\end{math}. For a sample of category $t$, the corresponding 
	class-balanced re-weighting loss $L_{CB}$ is formulated as
	\begin{equation} \label{CBloss}
		L_{CB} = -\omega_t \log{p_t}
	\end{equation}
	The derivative of the $L_{CB}$ with respect to the model’s class-k output $z_k$ is
	\begin{equation}
		\frac{\partial L_{CB}}{\partial z_k} = 
		\begin{cases}
			\omega_t(p_t-1), &\text{$k=t$} \\
			\omega_t p_k, &\text{$k\neq t$}
		\end{cases}
	\end{equation}
	Depending on the frequency of class $k$, the gradient contributions of 
	examples are far different. If category $k$ is a tail class, for example $k=C$, 
	in which case we have $\omega_k=\max_i \omega_i$, re-weighting is reasonable:
	\begin{enumerate}
		\item If $k=t$, a large encouraging gradient $\omega_t\left(p_t-1\right)$ is produced by a correct prediction for tail classes;
		\item 	If $k\neq t$, the discouraging gradient $\omega_tp_k$ is relatively small as $\omega_t<\omega_k$. 
		This is consistent with the idea of ignoring discouraging gradient for tail classes \cite{tan2020equalization}.
	\end{enumerate}
	However, when it comes to head classes, for example $k=1$, 
	in which case we have $\omega_k=\min_i \omega_i$, the learning process is seriously hindered:
	\begin{enumerate}
		\item If $k=t$, the encouraging gradient $\omega_t\left(p_t-1\right)$ 
		is very small as $\omega_t$ is close to zero; 
		\item If $k\neq t$, the discouraging gradient 
		$\omega_tp_k$ 
		is relatively large as $\omega_t>\omega_k$. 
		This suppression effect is further accumulated because each 
		positive sample of class k will be treated as a negative sample 
		for all other classes.
	\end{enumerate}
	In general, while tail classes benefit from re-weighting, the universal 
	representative ability is damaged as the dominant classes suffer 
	from overwhelmed discouraging gradients.
	
	\subsection{Motivation}
	
	As aforementioned, the re-weighting methods lead to sub-optimal result, because the focus 
	on tail classes is entangled with the overall representation learning 
	process in the classification loss. Accordingly, our key idea is to 
	decompose the task of long-tailed learning into two separate parts, 
	learning generalizable representations and facilitating learning for 
	tail classes. The motivation is two-fold.
	
	{\bfseries Motivation 1.} To learn the most generalizable representations, 
	directly re-weighting the cross-entropy loss should be carefully avoided. It has been 
	proved by experiments \cite{cao2019learning, zhou2020bbn, kang2019decoupling} 
	and our theoretical analyses in Sec. \ref{rw}.
	It motivates us to keep the instance-balanced cross-entropy loss unchanged, 
	which exhaustively takes advantage of the diversity of dominant data and
	guarantees the model to learn generalizable representations.
	
	{\bfseries Motivation 2.} To facilitate the learning process of tail classes, 
	we take advantage of the transfer ability of knowledge distillation. 
	The predictive distributions in knowledge distillation contain informative 
	dark knowledge which has low risk of overfitting to specific classes or 
	examples. However, if sharing the same architecture and data, 
	the student model is expected to exhibit similar performance to 
	the teacher model. If distilled directly from teacher model, the student
	model yields almost no performance gain. Instead of distilling all 
	the knowledge without distinction, we argue that the teacher should 
	place more emphasis on remedying shortcoming of student model, 
	which is the classification ability for tail classes.
	
	\section{Method}
	
	As motivated, we propose balanced knowledge distillation to decouple
	the two goals of long-tailed learning and achieve both simultaneously.
	In this section, we firstly revisit the conventional knowledge distillation method, 
	then describe the proposed method in detail. In the last, we further analyze the validity
	of the proposed method from the perspective of gradient.
	
	\subsection{Review of Knowledge Distillation}
	
	The conventional response-based knowledge distillation \cite{hinton2015distilling}
	consists of two steps. First, a teacher model is trained with 
	cross-entropy loss. Second, the student model is trained together with 
	ground truth targets in addition to the teacher’s soft targets. 
	
	Formally, as a supplement of notations in Section 3, we define the 
	network outputs of the teacher model as 
	$\hat{z}=\left[{\hat{z}}_1,...,{\hat{z}}_c\right]^T$ 
	and the class probability ${\hat{p}}_i$ is calculated as
	$\hat{p}_i=softmax(\frac{\hat{z}_i}{T})$, where $T$ is a temperature parameter 
	that controls the {\itshape softness} of probability distribution over classes. 
	For the convenience of analysis, we set $T=1$. Similarly, 
	with a slight abuse of the notation, 
	we re-define the student’s probability in a more general form, 
	$p_i=softmax\left(\frac{z_i}{T}\right)$.
	The loss for the student is a linear combination of the 
	cross-entropy loss $L_{CE}$ and a Kullback-Leibler divergence loss $L_{KL}$:
	\begin{align} \label{KD}
		L_{KD}&=\alpha L_{CE}+\left(1-\alpha\right)L_{KL}, \\
		where\ L_{KL}&=T^2\sum_{i}{{\hat{p}}_i \log\frac{{\hat{p}}_i}{p_i}},
	\end{align}
	where $\alpha$ is a hyperparameter controlling the trade-off between
	the two losses.
	
	\subsection{Balanced Knowledge Distillation}
	
	Although effective for model compression,
	the conventional knowledge distillation framework fails to improve the model performance
	dramatically
	by distilling another structured identically model. Particularly, if trained on 
	long-tailed data, the teacher model is naturally biased towards
	the head classes. In the distillation process, the predictive information for the
	tail classes is overwhelmed by the head classes. Therefore,
	the student model guided from such biased model may exhibit
	an even worse performance.
	
	To solve this problem, we propose 
	balanced knowledge distillation, which transfers knowledge with focus.
	Our BKD follows the teacher-student learning 
	pipeline as mentioned above. The key difference is that we take 
	class priors into consideration and control the importance
	of distilled information for different classes. Concretely, 
	given a teacher model trained 
	with vanilla cross-entropy loss, the student model is trained by 
	minimizing the summation of an instance-balanced cross-entropy loss 
	and a class-balanced distillation loss. The total loss of balanced 
	knowledge distillation is formulated as
	\begin{equation}
		L=L_{CE}+T^2\sum_{i}{\omega_i{\hat{p}}_ilog\frac{{\hat{p}}_i}{p_i}}.
	\end{equation}
	The weight factor $\omega_i$ is defined as Eq. \ref{class-balanced term}. 
	In this way, the dark knowledge 
	from tail classes is distilled with focus for facilitating learning on tail classes.
	
	Despite the validity from the perspective of distillation 
	with focus, the nonnegativity of KL-divergence is damaged 
	because the weighted probabilities of teacher model do not sum 
	to one any more. To keep the divergence loss nonnegative, we consider 
	$\omega^T\hat{p}$ as a whole and normalize it to one. Accordingly, 
	the loss can be rewritten as
	\begin{equation}\label{BKDEqu}
		L_{BKD}=L_{CE}+T^2\sum_{i}{\omega_i{\hat{p}}_ilog\frac{\omega_i{\hat{p}}_i}{p_i}}.
	\end{equation}
	Note that this modification has no impact on optimization as 
	the predictive probabilities of the teacher models contribute 
	zero gradient to the student model. With the normalization, a definite
	lower bound of the loss function is now guaranteed.
	
	Our BKD framework is summarized in Algorithm \ref{BKDAlg}.
	
	\subsection{Analysis of Gradient}
	
	To better understand the balanced knowledge distillation loss, we make a further 
	analysis from the perspective of gradient. The derivative of the 
	$L_{BKD}$ with respect to student model’s output $z$ is
	\begin{equation}
		\frac{\partial L_{BKD}}{\partial z_k}=-\left(\omega_k{\hat{p}}_k+y_k\right)+\sum_{i=1}^{C}\left(\omega_i{\hat{p}}_i+y_i\right)p_k,
	\end{equation}
	where the target to mimic for student model is $\omega_k\hat{p}_k+y_k$. 
	For the convenience of analysis, we normalize this term to sum to one, i.e.,  $\sum_{i=1}^{C}\left(\omega_i\hat{p}_i+y_i\right)=1$. Thus we have
	\begin{equation}
		\frac{\partial L_{BKD}}{\partial z_k} =
		\begin{cases}
			p_t-(\omega_t\hat{p}_t+1), &\text{$k=t$} \\
			p_k-\omega_k\hat{p}_k, &\text{$k\neq t$}
		\end{cases}
	\end{equation}
	For a head class $k$, $\omega_k \to 0$. The gradient gap between 
	balanced knowledge distillation loss and cross-entropy loss is negligible. Therefore,
	the representation learning process for the dominant classes is almost unaffected.
	For a tail class $k$, the target consists of both the ground truth target and
	the teacher's soft target. The informative predictive knowledge facilitates learning
	for the tail classes.
	
	\begin{algorithm}  [t]
		\caption{Balanced Knowledge Distillation}   
		\label{BKDAlg}  
		\begin{algorithmic}  [1]
			\REQUIRE A long-tailed dataset $D$, maximum epoch number $E$
			and weight $\omega$, as well as a teacher model $t_{\hat{\theta}}$
			and a student model $s_\theta$ with identical structure.
			\STATE Initialize parameters $\hat{\theta}$ randomly;
			\FOR{$e=1,\dots ,E$}
			
			\STATE Sample a minibatch $\mathcal{B}$ from $D$;
			\STATE Update $t_{\hat{\theta}}$ by minimizing $L_{CE}$
			(Equation \ref{CELoss}) on $\mathcal{B}$;
			
			\ENDFOR
			\STATE Initialize parameters $\theta$ randomly;
			\FOR{$e=1,\dots ,E$}
			
			\STATE Sample a minibatch $\mathcal{B}$ from $D$;
			\STATE Use $t_{\hat{\theta}}$ to produce predictions $\hat{p}$ on $\mathcal{B}$;
			\STATE Normalize $\omega^T \hat{p}$. 
			\STATE Update $s_\theta$ by minimizing $L_{BKD}$
			(Equation \ref{BKDEqu}) on $\mathcal{B}$;
			
			\ENDFOR
		\end{algorithmic} 
		
	\end{algorithm}
	
	From the above analysis, BKD is able to achieve the dual goals of learning generalizable
	representations and facilitating learning for tail classes.
	
	\section{Experiments}
	
	\subsection{Datasets}
	
	We evaluate the proposed method on five long-tailed datasets, including
	Long-tailed CIFAR-10/-100 \cite{cui2019class}, Places-LT \cite{liu2019large}, 
	ImageNet-LT \cite{liu2019large} and iNaturalist 2018 \cite{van2018inaturalist}.
	
	\noindent
	{\bfseries Long-tailed CIFAR-10 and CIFAR-100.}\quad 
	The original version of CIFAR-10 and CIFAR-100 contains 
	60,000 images, 50,000 for training and 10,000 for validation
	with 10 and 100 classes, respectively. Following the prior work \cite{cui2019class, cao2019learning}, 
	we use the long-tailed
	version of both the CIFAR datasets by downsampling examples
	per class with different ratios. The imbalance ratio $\rho$ denotes 
	the ratio between the number of training examples between
	the most frequent class and the least frequent class. We use $\rho=10,50,100$ 
	in our experiments.
	
	\begin{table*}[htbp]
		\caption{Top-1 validation accuracy of ResNet-32 on long-tailed CIFAR-10 and CIFAR-100. \dag \
			indicates results reported in \cite{zhou2020bbn}. \ddag \ indicates result reported in
			\cite{xiang2020learning}.
			Best results are marked in bold.} 
		\label{CIFAR}
		\centering		
		\begin{tabular}{ccccccc}		
			\toprule		
			Dataset & \multicolumn{3}{c}{Long-tailed CIFAR-10} & \multicolumn{3}{c}{Long-tailed CIFAR-100} \\	
			\cmidrule(r){1-1} \cmidrule(r){2-4} \cmidrule(r){5-7}
			imbalance ratio &  100		& 50   & 10 &  100		& 50   & 10 \\  		
			\midrule		
			CE  &  70.36  &  74.81  &  86.39  &  38.32  &  43.85  &  55.71\\
			KD \cite{hinton2015distilling}  &  70.69  &  77.92  &  87.48  &  40.36  &  45.49  &  59.22\\
			CB \cite{cui2019class}  &  72.11  &  77.73  &  86.40  &  32.65  &  38.57  &  54.87\\
			LDAM-DRW\dag\ \cite{cao2019learning}  &  77.03  &  81.03  &  88.16  &  42.04  &  46.62  &  58.71\\
			BBN\dag\ \cite{zhou2020bbn}  &  79.82  &  82.18  &  88.32  &  42.56  &  47.02  &  59.12\\
			Logit adjustment loss \cite{menon2020long}  &  77.93  &  81.64  &  88.17  &  42.01  &  47.03  &  57.74\\
			LFME\ddag\ \cite{xiang2020learning}  &  -  &  -  &  -  &  42.30  &  -  &  -\\
			\midrule
			BKD  &  {\bfseries 81.72}  &  {\bfseries 83.81}&  {\bfseries 89.21}    &  {\bfseries 45.00}  &  {\bfseries 49.64}  &  {\bfseries 61.33}\\
			\bottomrule		
		\end{tabular}	
	\end{table*}
	
	\begin{table}[htbp]
		\caption{Top-1 accuracy on iNaturalist 2018. \dag \ denotes 
			results reported in the original paper. \ddag \ denotes results reported in \cite{zhou2020bbn}.}
		\label{iNaturalist}
		\centering
		\begin{tabular}{ccccc}
			\toprule
			Method & Many & Medium & Few & All\\
			\midrule
			CE  &  72.2  &  63.1  &  57.4  &  61.8\\
			KD \cite{hinton2015distilling}  &  {\bfseries 72.6}  &  63.8  &  57.4  &  62.2\\
			CB \cite{cui2019class}  &  53.4  &  54.8  &  53.2  &  54.0\\
			CB-Focal\dag\ \cite{cui2019class}  &  -  &  -  &  -  &  61.1\\
			LDAM-DRW\ddag\ \cite{cao2019learning}  &  -  &  -  &  -  &  64.6\\
			BBN\ddag\ \cite{zhou2020bbn}  &  -  &  -  &  -  &  66.3\\
			cRT\dag\ \cite{kang2019decoupling}  &  -  &  -  &  -  &  65.2\\
			
			\midrule
			BKD  &  67.1  &  {\bfseries 66.1}  &  {\bfseries 67.6}  & {\bfseries 66.8}\\
			\bottomrule
		\end{tabular}
	\end{table}
	
	\begin{table}[htbp]
		\caption{Top-1 accuracy on Places-LT. \dag \ denotes results
			reported in \cite{xiang2020learning}. \ddag \ denotes results
			reported in \cite{kang2019decoupling}.}
		\label{Places}
		\centering
		\begin{tabular}{ccccc}
			\toprule
			Method & Many & Medium & Few & All\\
			\midrule
			CE\ddag\  &  {\bfseries 45.7}  &  27.3  &  8.2  &  30.2\\
			KD \cite{hinton2015distilling}  &  {\bfseries 45.7}  &  28.1  &  9.0  &  30.7\\
			CB \cite{cui2019class}  &  36.5  &  29.7  &  9.2  &  28.2\\
			Focal\dag\ \cite{lin2017focal}  &  41.1  &  34.8  &  22.4  &  34.6\\
			OLTR\dag\ \cite{liu2019large}  &  44.7  &  37.0  &  25.3  &  35.9\\
			LFME\dag\ \cite{xiang2020learning}  &  38.4  &  {\bfseries 39.1}  &  21.7  &  35.2\\
			cRT\ddag\ \cite{kang2019decoupling}  &  42.0  &  37.6  &  24.9  &  36.7\\
			\midrule
			BKD  &  41.9  &  {\bfseries 39.1}  &  {\bfseries 30.0}  & {\bfseries 38.4}\\
			\bottomrule
		\end{tabular}
	\end{table}
	
	\noindent
	{\bfseries Places-LT.}\quad 
	Places365-Standard \cite{zhou2017places} is a large-scale image database for scene recognition, with more than
	1.8 million training images from 365 categories. We construct Places-LT by the same sampling
	strategy as \cite{liu2019large}, with the number of images per class ranging from 4980 to 5. 
	
	\noindent
	{\bfseries iNaturalist 2018.}\quad 
	The iNaturalist species classification dataset \cite{van2018inaturalist} is a large-scale real-world dataset. 
	The iNaturalist 2018 dataset contains 437,513 training images from 8142 classes,
	with an imbalance ratio of 500. For fair comparisons, we use the official 
	training and validation splits in our experiments.
	
	\noindent
	{\bfseries ImageNet-LT.}\quad 
	ImageNet-LT is constructed by sampling a subset of ImageNet-2012 \cite{russakovsky2015imagenet} following 
	the Pareto distribution with the power value $\alpha=6$. It has 115.8K images
	from 1000 categories, with the number of images per class ranging from 1280 to 5. 
	
	\subsection{Implementation Details}
	
	\begin{table}[htbp]
		\caption{Top-1 accuracy on ImageNet-LT. \dag \ denotes results
			reported in \cite{xiang2020learning}. * denotes our reproduced results with released code 
			from \cite{kang2019decoupling}.}
		\label{ImageNet}
		\centering
		\begin{tabular}{ccccc}
			\toprule
			Method & Many & Medium & Few & All\\
			\midrule
			CE*  &  56.9  &  25.4  &  3.6  &  34.6\\
			KD \cite{hinton2015distilling}  &  {\bfseries 58.8}  &  26.6  &  3.4  & 35.8\\
			CB \cite{cui2019class}  &  39.6  &  32.7  &  16.8  &  33.2\\
			Focal\dag\ \cite{lin2017focal}  &  36.4  &  29.9  &  16.0  &  30.5\\
			OLTR\dag\ \cite{liu2019large}  &  43.2  &  35.1  &  18.5  &  35.6\\
			LFME\dag\ \cite{xiang2020learning}  &  47.1  &  35.0  &  17.5  &  37.2\\
			cRT* \cite{kang2019decoupling}  &  51.5  &  38.3  &  {\bfseries 22.8}  &  41.2\\
			\midrule
			BKD  &  54.6  &  37.2  &  20.4  & 41.6\\
			BKD+cRT  &  52.3  &  {\bfseries 39.8}  &  22.4  & {\bfseries 42.3}\\
			\bottomrule
		\end{tabular}
	\end{table}
	
	\begin{figure*}[htbp]
		\centering
		\subfigure{
			\includegraphics[width=2.5in]{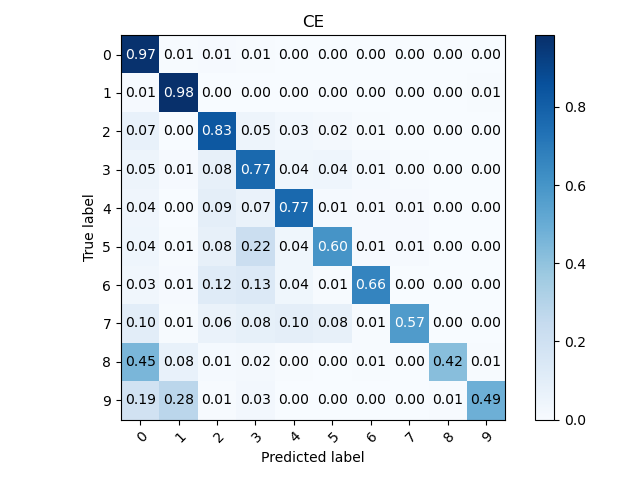}
		}
		\subfigure{
			\includegraphics[width=2.5in]{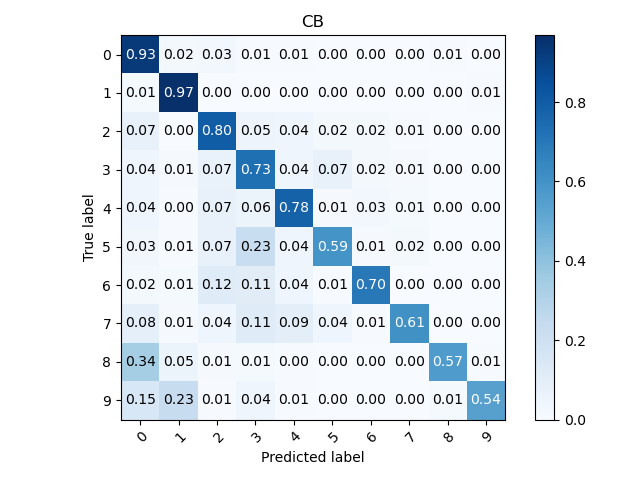}
		}
		\quad    
		\subfigure{
			\includegraphics[width=2.5in]{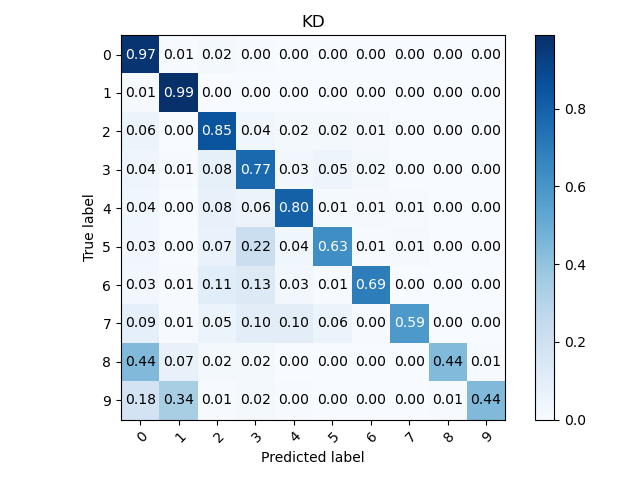}
		}
		\subfigure{
			\includegraphics[width=2.5in]{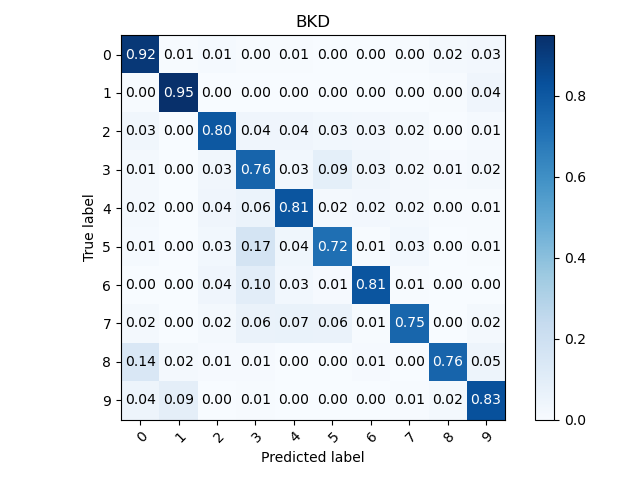}
		}
		\caption{Illustration of the confusion matrices by 
			the CE, CB, KD and our BKD on long-tailed CIFAR-10 ($\rho=100$)}
		\label{confusion}
	\end{figure*}
	
	In our experiments, the teacher model and 
	the student model have identical architecture on each dataset.
	As summarized in Algorithm \ref{BKDAlg}, we first train the teacher model with vanilla cross-entropy loss
	and then train the student model with the proposed BKD loss. 
	In all experiments, we set $\beta=0.9999$ in Equation \ref{class-balanced term} and
	the temperature $T=2$ in Equation \ref{BKDEqu}. 
	All networks are trained with SGD with a momentum of 0.9.
	Unless otherwise specificed, the base learning rate is set to 0.2, 
	with cosine learning rate decay.
	Other details are given below.
	
	\noindent
	{\bfseries Implementation details for long-tailed CIFAR datasets.}\quad 
	We employ ResNet-32 as the backbone 
	network and follow the training recipe of \cite{he2016deep} for the teacher model 
	and the student model. Both models are trained for 200 epochs 
	with the batch size of 128. The learning rate is initialized as 0.1 and
	decayed by 0.01 at the 160th epoch and again at the 180th epoch. For long-tailed
	CIFAR-10, we first train the student model with vanilla knowledge distillation
	before the 160th epoch, and then deploy our BKD, following \cite{cao2019learning}.
	
	\noindent
	{\bfseries Implementation details for Places-LT.}\quad 
	We choose pretrained ResNet-152 as the backbone network, following \cite{liu2019large}. Both the teacher model
	and the student model are trained sequentially for 90 epochs with the batch size of 128. 
	
	\noindent
	{\bfseries Implementation details for ImageNet-LT.}\quad 
	We train ResNet-10 from scratch for ImageNet-LT. Both the 
	models are trained sequentially for 90 epochs with the batch size of 512. 
	
	\noindent
	{\bfseries Implementation details for iNaturalist 2018.}\quad 
	We use ResNet-50 as our backbone network. Both the 
	models are trained sequentially for 90 epochs with the batch size of 256. 
	
	\subsection{Experimental Results}
	
	
	\begin{table*}[htbp]
		\caption{Accuracy summary of CE, CB, KD and our BKD. The best (worst)
			results are marked in green (red).}
		\label{ablation}
		\centering
		\begin{tabular}{cccccc}
			\toprule
			Method & Long-tailed CIFAR-10 & Long-tailed CIFAR-100 & iNaturalist 2018 & Places-LT & ImageNet-LT\\
			\midrule
			CE  &  \textcolor{red}{70.4}  &  38.3  &  61.8  &  30.2  &  34.6\\
			CB \cite{cui2019class}  &  72.1  &  \textcolor{red}{32.7}  &  \textcolor{red}{54.0}  &  \textcolor{red}{28.2}  &  \textcolor{red}{33.2}\\
			KD \cite{hinton2015distilling}  &  70.7  &  40.4  &  62.2  &  30.7  &  35.8\\
			\midrule
			BKD  &  \textcolor[rgb]{0,0.75,0}{81.7}  &  \textcolor[rgb]{0,0.75,0}{45.0}   &  \textcolor[rgb]{0,0.75,0}{66.8}   &
			\textcolor[rgb]{0,0.75,0}{38.4}   &  \textcolor[rgb]{0,0.75,0}{41.6} \\
			\bottomrule
		\end{tabular}
	\end{table*}
	
	
	\noindent
	{\bfseries Competing methods.}\quad 
	In experiments, we compare our proposed BKD with the standard training and several 
	state-of-the-art methods, including: 
	(1) {\itshape CE:} standard training with vanilla cross-entropy loss;
	(2) {\itshape KD:} conventional knowledge distillation technique \cite{hinton2015distilling} 
	by minimizing Equation \ref{KD}; 
	(3) {\itshape CB:} re-weighting the loss according to the inverse of the effective number of samples 
	in each class \cite{cui2019class}, defined as Equation \ref{CBloss}. It can also be integrated with focal loss\cite{lin2017focal}, denoted CB-Focal; 
	(4) {\itshape Focal:} focal loss \cite{lin2017focal}; 
	(5) {\itshape LDAM-DRW:} the integration of LDAM loss and deferred re-weighting strategy \cite{cao2019learning}; 
	(6) {\itshape LFME:} learning from multiple experts with adaptive learning schedules \cite{xiang2020learning}; 
	(7) {\itshape OLTR:} Open Long-Tailed Recognition \cite{liu2019large}.
	(8) {\itshape BBN:} Bilateral-Branch Network \cite{zhou2020bbn}; 
	(9) {\itshape cRT:} first learning representation and then re-training classifier \cite{kang2019decoupling};
	(10) {\itshape Logit Adjustment loss:} the logit adjusted softmax cross-entropy loss \cite{menon2020long}.
	
	\begin{figure}[t]
		\centering
		\includegraphics[width=\linewidth]{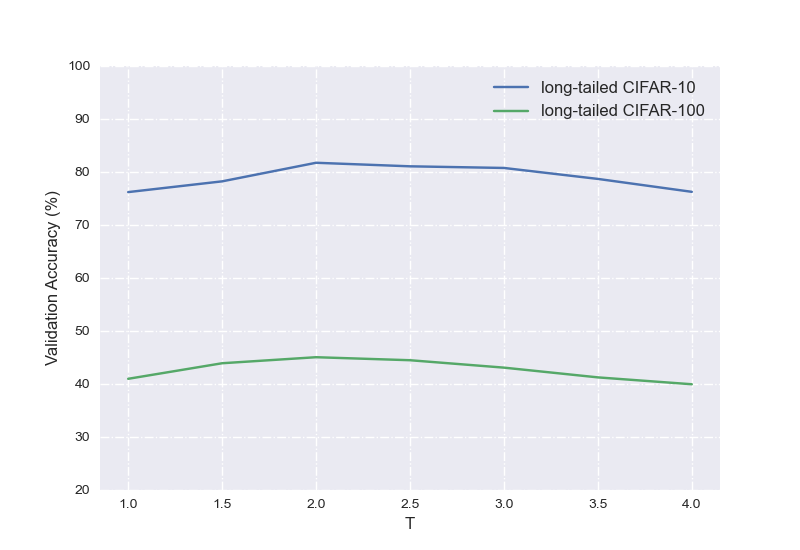}
		\caption{Comparison of the top-1 validation accuracy when varying temperature parameter $T$ on 
			long-tailed CIFAR datasets.}
		\label{tem}
	\end{figure}
	
	\begin{figure}[t]
		\centering
		\includegraphics[width=\linewidth]{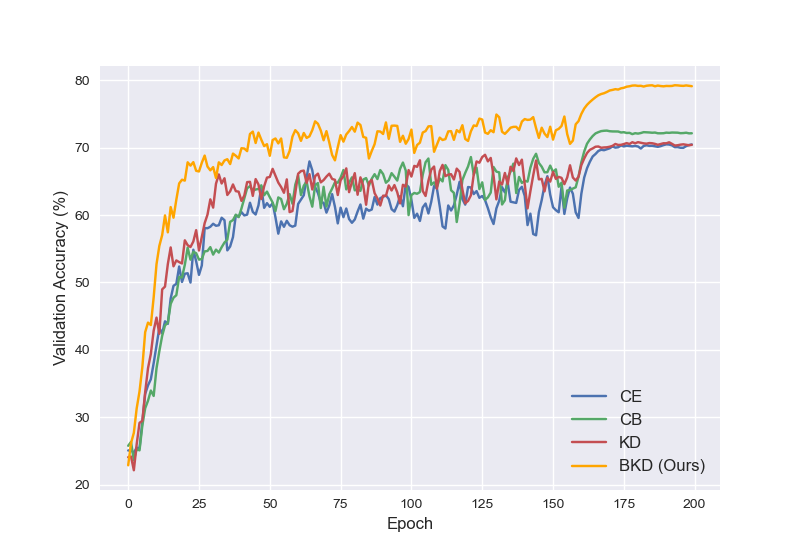}
		\caption{Illustration of the accuracy curves by the CE, CB, KD and our BKD.}
		\label{curves}
	\end{figure}
	
	\noindent
	{\bfseries Results on long-tailed CIFAR datasets.}\quad 
	We conduct experiments on long-tailed CIFAR-10/-100 with
	three different imbalance ratios $\rho = 10,50, 100$. 
	Table \ref{CIFAR} shows the validation accuracy of various methods
	for long-tailed CIFAR datasets. 
	It can be seen that the proposed BKD significantly outperforms
	the cross-entropy and knowledge distillation \cite{hinton2015distilling}. 
	It also demonstrates that the proposed BKD exhibits superior
	performance compared with existing state-of-the-art methods,
	including recently proposed BBN \cite{zhou2020bbn} and the logit adjustment loss \cite{menon2020long} method.
	
	\noindent
	{\bfseries Results on large-scale datasets.}\quad 
	We validate the effectiveness of our method on three large-scale datasets. 
	Besides the overall top-1 classification accuracy, we also calculate the 
	accuracy of three subsets: {\itshape Many-shot classes} 
	(with over 100 training samples), {\itshape Medium-shot classes} (with $20\sim100$ training samples)
	and {\itshape Few-shot classes} (with under 20 training samples).
	Table \ref{iNaturalist} summarizes the top-1 accuracy for iNaturalist 2018. From the
	table we  can see that our BKD outperforms the cross-entropy baseline as well as the previous
	state-of-the-art methods, including BBN \cite{zhou2020bbn} and cRT \cite{kang2019decoupling}. 
	It is worth noting that the performance on 
	tail classes (denoted as {\itshape Few} in the table) is significantly improved by our method and the performance gap
	between each subset thus becomes negligible. For Places-LT, we observe a similar improvement. As shown in 
	Table \ref{Places}, compared to the state-of-the-art methods,
	our BKD achieves the best overall accuracy, with more than
	$4.7\%$ performance gain on the {\itshape Few-shot} subset.
	We further evaluate the proposed BKD on the ImageNet-LT.
	From Table \ref{ImageNet} we observe that the results are 
	consistent with other datasets. It is also noteworthy that our 
	BKD outperforms the LFME \cite{xiang2020learning} by a large margin, even without
	multiple teacher models and complex learning schedules.
	Moreover, we test the combination of cRT \cite{kang2019decoupling} and our BKD, where 
	we first train the student model with BKD and then retrain 
	the classifier with class-balanced sampling.
	This combination further improves the classification
	performance on ImageNet-LT.
	
	\noindent
	{\bfseries The effect of the temperature.}\quad 
	In Figure \ref{tem}, we study the effect of tuning the temperature parameter $T$ ranging from $1.0$ to $4.0$ on the long-tailed CIFAR
	datasets. We find the optimal temperature is $2.0$ for both the datasets. Accordingly, we fix $T=2$ in all 
	our experiments.
	
	\subsection{Comparisons with KD and CB}
	
	In this section, we emphatically compare our BKD with the two most relevant methods, 
	i.e., knowledge distillation \cite{hinton2015distilling} and class-balanced re-weighting \cite{cui2019class}. 
	
	
	First, In Figure \ref{confusion} we visualize four 
	confusion matrices respectively by the models of the cross-entropy training (CE), class-balanced
	re-weighting (CB), knowledge distillation (KD) and BKD on long-tailed CIFAR-10. The improvement from knowledge distillation and re-weighting is marginal, while BKD significantly improves the performance, 
	especially for tail classes.
	
	Second, we reorganize some of the results into Table \ref{ablation} 
	for convenient comparison. As we can see, KD is slightly superior to vanilla
	cross-entropy training due to the ability 
	of transferring knowledge. However, class-balanced
	re-weighting method is only effective on Long-tailed CIFAR-10 dataset. 
	When dealing with large-scale and extremely imbalanced datasets, 
	CB leads to even worse performance.
	The phenomena is consistent with our analysis in Section \ref{rw}. 
	Compared to the conventional KD, our BKD consistently outperforms 
	by $4.6\%\sim 11.0\%$ on the reported datasets.
	Moreover, we also visualize the accuracy
	curves during training for the four methods in Figure \ref{curves}. We find that BKD leads to higher validation accuracy
	throughout the training process.
	
	From the above comparisons, we can conclude that our BKD is an effective knowledge distillation framework
	in long-tailed scenarios, as well as a new state-of-the-art method for long-tailed learning.
	
	\section{Conclusion}
	
	In this work, we explore knowledge distillation 
	in long-tailed scenarios and propose a novel balanced
	knowledge distillation framework. We analyze
	and experimentally demonstrate that our BKD framework 
	effectively distills the knowledge on imbalanced 
	data by learning generalizable representations and facilitating learning 
	for tail classes simultaneously. Experimental results on several long-tailed benchmarks
	show that our BKD achieves state-of-the-art performance .

	{\small
		\bibliographystyle{ieee_fullname}
		\bibliography{arxiv.bib}
	}
	
\end{document}